# PGDP5K: A Diagram Parsing Dataset for Plane Geometry Problems

Yihan Hao[1,2], Mingliang Zhang[2,3], Fei Yin[2,3], Lin-Lin Huang[1]

[1]School of Electronic Information Engineering, Beijing Jiaotong University, Beijing, China
[2]National Laboratory of Pattern Recognition (NLPR), Institute of Automation of Chinese Academy of Science, Beijing, China
[3]School of Artificial Intelligence, University of Chinese Academy of Sciences, Beijing, China

*Abstract*—Diagram parsing is an important foundation for geometry problem solving, attracting increasing attention in the field of intelligent education and document image understanding. Due to the complex layout and between-primitive relationship, plane geometry diagram parsing (PGDP) is still a challenging task deserving further research and exploration. An appropriate dataset is critical for the research of PGDP. Although some datasets with rough annotations have been proposed to solve geometric problems, they are either small in scale or not publicly available. The rough annotations also make them not very useful. Thus, we propose a new large-scale geometry diagram dataset named PGDP5K and a novel annotation method. Our dataset consists of 5000 diagram samples composed of 16 shapes, covering 5 positional relations, 22 symbol types and 6 text types. Different from previous datasets, our PGDP5K dataset is labeled with more fine-grained annotations at primitive level, including primitive classes, locations and relationships. What is more, combined with above annotations and geometric prior knowledge, it can generate intelligible geometric propositions automatically and uniquely. We performed experiments on PGDP5K and IMP-Geometry3K datasets reveal that the state-of-the-art (SOTA) method achieves only 66.07% F1 value. This shows that PGDP5K presents a challenge for future research. Our dataset is available at http://www.nlpr.ia.ac.cn/databases/CASIA-PGDP5K/.

## I. INTRODUCTION

Geometry is the basic content of mathematics, which studies the structure and properties in the space. Solving geometry problems is a compulsory course for cultivating abstract thinking and spatial perception ability of students in the primary and secondary education. In recent years, geometry problem solving has gained much attention for its important applications in intelligent education community [1][2][3]. Given the plane geometry diagram and corresponding problem text, problem solving involves three extremely complicated processes: parsing the diagram and text [4][5], reasoning the geometric logic, and finally calculating the answer. Among them, the diagram parsing (Figure 1) involves primitive extraction and relation reasoning, regarded as one of foundational steps [6][7], to give the geometric description including detailed primitives and relationships of diagrams. The performance of diagram parsing has a significant impact on whether the geometry could be solved correctly or not.

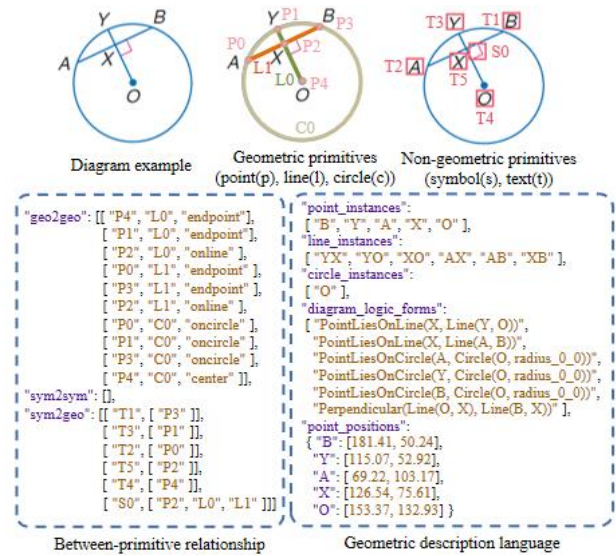

Fig. 1. An annotation example in PGDP5K dataset.

An appropriate dataset is critical to the geometry diagram parsing research and facilitate explorations of various high-performance algorithms. Although a few datasets [8][9] have been proposed recently, they have small number of samples, annotation errors or low-quality diagram images. Also, the layouts and styles of geometries, symbols and text in their samples are relatively simple, inconsistent with real samples in textbooks or exercise sets. Moreover, these datasets are only designed for traditional methods [1][10] such as Hough Transform with coarse-grained annotation, which makes them unsuitable for development of new algorithms in the deep learning era.

To promote the research of plane geometry diagram parsing (PGDP), this paper proposes a new large-scale diagram parsing dataset PGDP5K, consisting of 5000 diagram samples composed of 16 shapes, covering 5 positional relations, 22 symbol types and 6 text types. The various primitive types and complicated diagram layout of our dataset significantly reflect the challenge in diagram parsing task. Besides, we also propose a novel primitive-level annotation approach. The annotation contents include primitive class, parsing position, text OCR content, and between-primitive relation. The dense and accurate labeling provides a strong supervision for all subtasks of diagram parsing and makes the application of deep learning methods possible.

TABLE I. COMPARISON OF PGDP5K WITH EXISTING DATASETS

| Dataset | Number | Shape | Grade | Polygon Images | Annotation |
|---|---|---|---|---|---|
| GeoShader | 102 | 4 | 6-10 | / | / |
| GEOS | 186 | 4 | 6-10 | / | / |
| GEOS++ | 1406 | 4 | 6-10 | / | / |
| GEOS-OS | 2235 | 4 | 6-10 | / | / |
| Geometry3K | 3002 | 6 | 6-12 | 86 | Coarse-grained |
| PGDP5K | 5000 | 13 | 6-12 | 304 | Fine-grained |

Figure 1 shows an example of geometry diagram in our dataset with annotations of geometric primitives (points, lines and circles) and non-geometric primitives (symbols and texts). We also annotate the relations among primitives, which are critical but not available in existing datasets. The between-primitive relations consist of the relation between two or more geometric primitives and non-geometric primitives. These geometric relations, as the important cues of people's cognition of the diagram, are either explicit or hidden in the diagram. Furthermore, with the one-to-one correspondence between the relation and geometric propositions, we can generate the geometric description language (GDL) automatically and uniquely. The GDL, integrated with the basic information about point positions, geometric primitive instances and diagram logic forms, summarizes the contents of figures at high level. And so, it can facilitate the downstream applications, e.g., geometry problem solving [5][6][8][10][11][12] and geometry figure searching [13][14][15].

To assess the task difficulty, we conducted experiment with the existing state-of-the-art methods on the PGDP5K dataset as well as existing datasets Geometry3K and IMP-Geometry3K (re-annotated dataset of Geometry3K using our annotation method). The results show that the re-annotated dataset IMP-Geometry3K reports significantly higher accuracy than the existing dataset Geometry3K. This indicates that our fine-grained annotation method plays an important role in improving the ability of diagram parsing. Nevertheless, the best method only achieves 66.07% F1 value on our PGDP5K, indicating that our dataset presents a challenge for the future research.

## II. RELATED WORK

Geometry diagram parsing is a basic task of geometric problem solving and has been attracting much attention in the intelligent education community. Several datasets for geometry problem solving have been released in recent years. Although these datasets contain geometric diagrams and corresponding text contents, they have no detailed annotations and possess few geometry types. To our knowledge, there was no dataset offering detailed annotations for geometry diagram parsing.

Geometric problem solving datasets include GEOS [1], GEOS-OS [2], GeoQA [8], GeoShader [9], Geometry3K [10] and GEOS++ [16]. These datasets are relatively small in scale and contain limited geometry types. For example, there are only 102 shaded area problems in GeoShader. The GEOS dataset has only 186 problems, containing corresponding diagrams with simple layout. The GeoQA dataset contains 5,010 geometric problems, but its diagrams are unlabeled and of low quality. The GEOS++ and GEOS-OS datasets contain more samples of 1,406 and 2,235 problems, respectively, but they have not been publicly available yet.

The recently proposed Geometry3K dataset contains 3002 geometric problems with annotations of problem text and diagram. It covers not only four elementary shapes (lines, circles, regular quadrilaterals, and triangles) mentioned in GEOS, but also irregular quadrilaterals and other polygons. However, there are many duplicate diagram samples and annotation errors in Geometry3K. Additionally, it only gives coarse-grained annotations of diagrams, with neither detailed classification for different types of texts and symbols, nor relationships among primitives. These shortages limit the extraction of key information in geometric diagrams, and leads to low accuracy of geometric problem solving.

We outline the basic information of the current geometry diagram datasets in Table I. Compared with the existing datasets, our dataset is a large-scale and fine-grained annotated plane geometry diagram parsing dataset. In addition to lines, circles, triangles and regular quadrilaterals, it also contains a considerable number of irregular polygon images. Furthermore, unlike the coarse-grained annotated Geometry3K, PGDP5K has fine-grained primitive annotations and between-primitive relation annotations. These annotations make the application of deep learning methods possible in the field of geometry diagram parsing.

## III. DATABASE CONSTRUCTION

The collection of all diagrams and the annotation of primitives and relations were completed by four students in our group, spending a total of 500 hours. We developed our dataset in five steps: A. Dataset Collection and Classification; B. Primitive Annotation; C. Relation Annotation; D. Geometric Description Language Generation; E. Reviewing and Checking.

### A. Dataset Collection and Classification

Considering that there are a lot of duplicate diagram samples in Geometry3K dataset, we firstly checked duplicate of the 3002 diagram samples. After removing the duplicate samples, 1813 images were obtained. Then we collected other 3187 non-duplicated and diverse images from three popular textbooks across grades 6-12 on mathematics curriculum website[1].

We classified the collected images according to different geometric shapes. The geometric shapes of PGDP5K dataset are subject to long-tailed distribution as shown in Figure 2 (a). It is shown that about 41% of images contain triangles, about 24% contain circles and about 6% contain polygons. This shows that the proposed dataset not only conforms to the distribution law of textbooks from primary and secondary school, but also has geometric shape diversity.

### B. Primitive Annotation

For collected geometry images, we labeled geometric primitives and non-geometric primitives (symbols and texts) in diagram. According to primitive types, we grouped them as follows (Figures 5-7 display several instances of each class).

---

[1] https://www.mheducation.com/

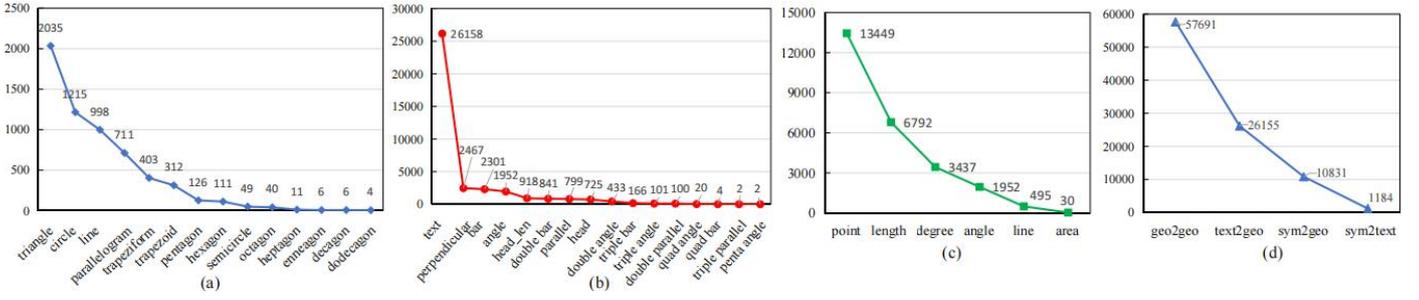

Fig. 2. Distribution of PGDP5K Dataset. (a)(b)(c)(d) respectively denote class distribution of geometry shape, symbol, text and relation.

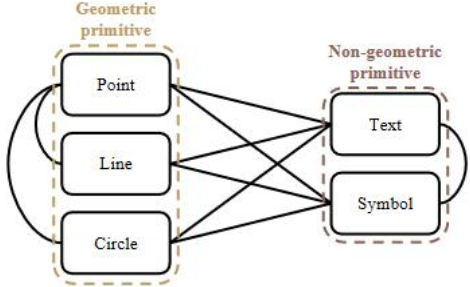

Fig. 3. Primitive relationship graph of plane geometry diagram.

*1) Geometric primitives:* We divided geometric primitives into 3 classes: point, line and circle. The point covers intersection point, tangent point, endpoint and independent point. The line consists of solid line, dash line and mixture of solid and dash. It is worth noting that we only label the longest line segment of all collinear lines. The circle includes complete circle and arc.

*2) Symbols:* We divided symbols into 6 super-classes and 16 sub-classes: perpendicular, angle, bar, parallel, arrow and head, where classes of angle, bar and parallel have multiple forms. As shown in Figure 4, the indicating arrows are combined with thick heads and spindly tails. The Geometry3K has a rough annotation of arrows, resulting in unsatisfactory performance of indication relation matching. Therefore, we annotated the arrows in more detail, not only the whole arrow, but also the head area of arrow. We subdivided the heads into two classes to distinguish different indication relations of different arrows. The class distribution of symbols in our dataset is shown in Figure 2 (b), where the text is seen as a special symbol class.

*3) Texts:* We divided texts into 6 classes. Geometry3K dataset classifies all texts into one rough text category. In many cases, there is no visual and content distinction between different text classes, e.g., angle, length and degree. Therefore, coarse-grained text annotation may confuse the subsequent geometric relation extraction and be not conducive to the task of diagram parsing. Considering that we divided the texts into line, point, angle, length, degree and area, making fine-grained text classification as a new sub-task of diagram parsing. The class distribution of texts is displayed in Figure 2 (c).

As to geometric primitives, we annotated their parsing positions and uniform pixel widths with the annotation tool

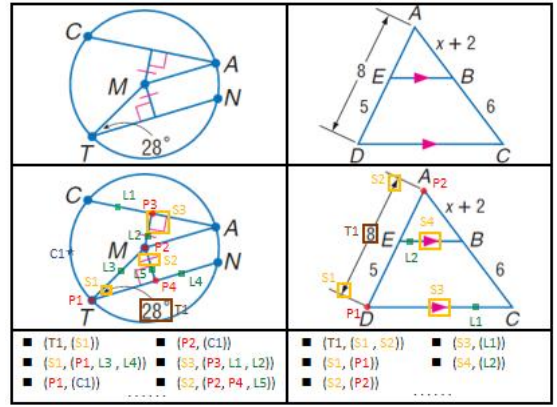

Fig. 4. Relation tuples of PGDP5K dataset. 'P#', 'L#', 'C#', 'T#' and 'S#' denote instances of point, line, circle, text and symbol, respectively.

Labelme[2]. For non-geometric primitives, we annotated the bounding box, symbol class and text class with the annotation tool LabelImg[3], and recorded corresponding text contents. To reduce the workload of manual labeling, we used a semi-automatic generation method to annotate our database. Firstly, we manually annotated 1500 images to roughly find out the general features of lines, circles and various types of symbols. Then, we used the Hough Transform algorithm [17] to extract geometric primitives and adopted the existing universal text detector to detect non-geometric primitives. Finally, we manually modified pre-generated annotations in the annotation tools to improve the accuracy of annotations.

### C. Relation Annotation

As to primitive relations, we constructed a relation graph of most basic relations among primitives in Figure 3. We divided primitive relations into 4 classes: geo2geo, text2geo, sym2geo and sym2text. For relations of geometric primitives, we only construct relations between point and line, point and circle, because other high-level relations among geometric primitives could be exported from these two basic relations. Inspires from the scene graph generation (SGG) problem [18][19], we defined a two-tuple with multiple entities to represent the relation between primitives. We take points, symbols and texts as subjects, and set other primitives related as objects. Some relation tuples are shown in Figure 4. Take the fourth two-tuple (S3, (P3, L1, L2)) of the left image as an example, where S3 denotes the perpendicular symbol, P3, L1 and L2 represent geometric primitives. It indicates that the symbol S3 acts on the angle formed by lines L1 and L2 at point P3, which means that L1 is perpendicular to L2. Take the first two-tuple (T1, (S1, S2)) of the right image as another example, where T1 denotes the length text, S1 and S2 represent head_len symbols. This tuple indicates that the text T3 acts between the symbols S1 and S2, that is, the length between S1 and S2 is T1.

---

[2] http://labelme.csail.mit.edu/Release3.0/
[3] https://github.com/tzutalin/labelImg

TABLE II. GEOMETRIC PROPOSITION TEMPLATES OF PRIMITIVE RELATION. "$" REPRESENTS GEOMETRIC PRIMITIVES AND "&" DENOTES TEXT CONTENT.

| Relation Class | Primitive Class | Proposition Templates |
|---|---|---|
| Geo Shape | Point | · Point($) |
| | Line | · Line($,$), Line($) |
| | Circle | · Circle($,radius_$) |
| | Angle | · Angle($,$,$), Angle($) |
| | Arc | · Arc($,$), Arc($,$,$) |
| Geo2Geo | Point | · PointLiesOnLine($,Line($,$))<br>· PointLiesOnCircle($,Circle($,radius_$))<br>· Circle($,radius_$) |
| Text(&)2 Geo | Text_point | · Point(&) |
| | Text_line | · Line(&) |
| | Text_angle | · Equals(MeasureOf(Angle($,$,$)),MeasureOf(angle &)) |
| | Text_degree | · Equals(MeasureOf(Angle($,$,$)), &)<br>· Equals(MeasureOf(Arc($,$)), &) |
| | Text_length | · Equals(LengthOf(Line($,$)), &)<br>· Equals(LengthOf(Arc($,$)), &) |
| | Text_area | - |
| Sym2 Geo | Sym_perpendicular | · Perpendicular(Line($,$), Line($,$)) |
| | Sym_angle | · Equals(MeasureOf(Angle($,$,$)), MeasureOf(Angle($,$,$))) |
| | Sym_bar | · Equals(LengthOf(Line($,$)), LengthOf(Line($,$)))<br>· Equals(LengthOf(Arc($,$)), LengthOf(Arc($,$))) |
| | Sym_parallel | · Parallel(Line($,$), Line($,$)) |

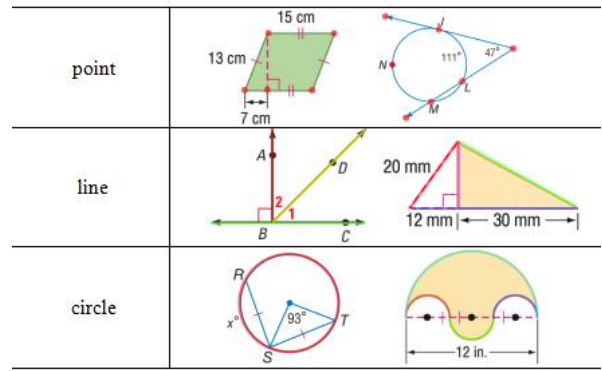

Fig. 5. Examples of geometric primitive.

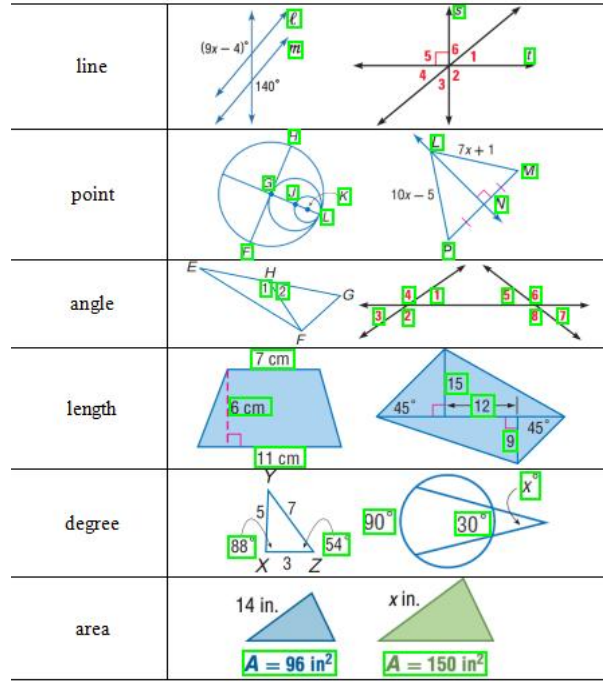

Fig. 6. Examples of text.

We observe that the relation extraction needs to be combined with complex visual, semantic, spatial and structural information, whereas quite a few of simple relations could be derived from ordinary rules of text content and space location. Therefore, we first employed the distance-based and content-based rules [10] to construct coarse relationships, then assigned the primitive relations generated to the annotator for correction. The class distribution of primitive relation is described in Figure 2 (d).

*D. Geometric Description Language Generation*

We formed the high-level and comprehensible specifications of GDL, which mainly consists of a list of geometric propositions formatted by proposition templates. In our annotation way, the GDL can be automatically generated according to the annotated relations and primitive classes due to one-to-one correspondences between primitive relations and geometric propositions. Besides, as shown in Table II, we defined four types of proposition templates about basic relations: Geometry Shape, Geo2Geo, Text2Geo and Sym2Geo.

- Geometry shapes are basic elements of high-level propositions. We give proposition templates of five types of fundamental geometry shapes: point, line, circle, angle and arc, where line, angle and arc have several equivalent expressions.

- Geo2Geo: three types of proposition templates are defined for relations among geometric primitives: point lies on line, point lies on circle and point is center of circle. These three primary relations could produce more other high-level relations of geometric primitives.

- Text2Geo: the relations of text with geometric primitives are divided into six types according to text class. Among the six proposition templates, the ones of degree and length are not unique.

- Sym2Geo: similar to the relations of text with geometric primitive, the propositions of symbol with geometric primitive are divided into four groups according to symbol class, and there are two proposition templates of symbol bar.

It deserves mentioning that the proposition templates are the crystallization of geometry knowledge, because they not only formulate the primitive relations but also facilitate the symbolic reasoning of problem solving.

| | |
|---|---|
| perpendicular | |
| angle<br>double angle<br>triple angle<br>quad angle<br>penta angle | |
| bar<br>double bar<br>triple bar<br>quad bar | |
| parallel<br>double parallel<br>triple parallel | |
| arrow | |
| head | |
| head_len | |

Fig. 7. Examples of symbol.

*E. Reviewing and Checking*

Reviewing the entire annotation process, we first labeled primitives, then annotated the relations between primitives, and finally generated the GDL according to the one-to-one correspondence between primitive relations and geometric propositions. Hence, annotations of primitives and relations determine the correctness of the subsequent description language generation.

According to annotation process, we designed the checking process demonstrated in Figure 8 to inspect annotations. To help check existing annotation errors, we first visualized the annotated primitives and relations on the original images. Then the annotation situation was displayed to help modify all errors in both primitive and primitive relation annotation. Finally, we regenerated the DGL and restarted another loop of check process until no error exists.

*F. Dataset Statistics*

Our dataset PGDP5K has a total of 5000 images, and is divided into the training set, validation set, and test set at the ratio of 0.7:0.1:0.2. It covers 16 shapes, 5 positional relations, 16 symbol types and 6 text types. Figure 2 shows the distributions of geometric shape, symbol, text and relation class. Obviously, the proposed dataset is subject to long-tailed distribution.

To sum up, the diagrams in PGDP5K have more complicated layouts, and even primitives of same class have great difference in style. This makes our dataset more challeng-ing. Meanwhile, PGDP5K dataset provides more detailed annotations at primitive level so as to facilitate tasks of diagram parsing and geometric problem solving.

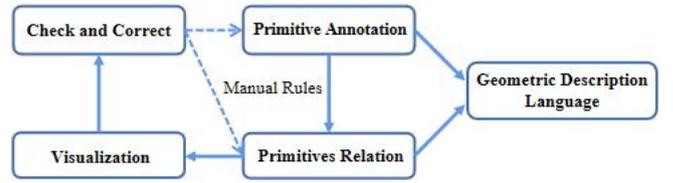

Fig. 8. Check process of annotation.

## IV. EXPERIMENTS

*A. Parsing Methods*

To assess the difficulty of our dataset, we employed three diagram parsing methods to evaluate the performance in two tasks: geometric primitive detection and GDL generation. The three methods are below.

- Freeman [20]: Considering the characteristics of Freeman chaincode of circle and line, the corresponding properties are obtained, so as to realize the detection of geometric primitives.

- GEOS [1]: This is a geometry problem solving system, whose geometric primitive detection method is based on Hough Transform [17]. The detection of points, lines and circles is completed through five modules: image segment parsing, primitive parsing, primitive selecting, core parsing and graph parsing.

- Inter-GPS [10]: This is a latest state-of-the-art model in the field of geometric problem solving. Its geometric primitive extraction module is based on the GEOS [1]. The symbol and text regions in diagram are extracted through a strong object detector RetinaNet [21], and the text contents are recognized by the optical character recognition (OCR) tool MathPix. After obtaining the primitive set and symbol set, it grounds each primitive with its associated primitives by distance-based and content-based rules.

*B. Performance Metrics*

*1) Geometric primitives:* An extracted geometric primitive is judged to be correct or not according to the distance evaluation of parsing position. We set distance threshold as 15 consistent with the Inter-GPS. Experimental results are evaluated in three metrics: Precision (P), Recall (R) and F1 score (F1).

*2) Geometric description language:* The logic forms of the geometric description are characterized by diversity and equivalence, for example, "Angle(C,D,E)" is equivalent to "Angle(C,D,A)" in the right panel of Figure 4. For rationality and fairness of evaluation, we improve the existing evaluation method focusing on propositions about line and angle. Experimental results are evaluated in seven metrics: P, R, F1, Likely Same (LS, F1≥50%), Almost Same (AS, F1≥75%), Perfect Recall (PR, recall=100%) and Totally Same (TS, F1=100%). More details of these metrics can be found in [10].

TABLE III. DETECTION PERFORMANCE OF GEOMETRIC PRIMITIVE

|  |  | Freeman | | GEOS | |
|---|---|---|---|---|---|
|  |  | *Geometry3K* | *PGDP5K* | *Geometry3K* | *PGDP5K* |
| Point | P | 72.56 | 68.23 | 81.91 | 79.94 |
|  | R | 80.33 | 81.31 | 90.97 | 93.34 |
|  | F1 | 76.25 | 74.20 | 86.21 | 86.13 |
| Line | P | 61.93 | 52.65 | 72.84 | 66.77 |
|  | R | 78.62 | 80.69 | 88.28 | 90.52 |
|  | F1 | 69.29 | 63.72 | 79.82 | 76.85 |
| Circle | P | 91.84 | 90.76 | 98.61 | 98.25 |
|  | R | 96.92 | 97.69 | 98.89 | 99.24 |
|  | F1 | 94.31 | 94.10 | 98.75 | 98.74 |

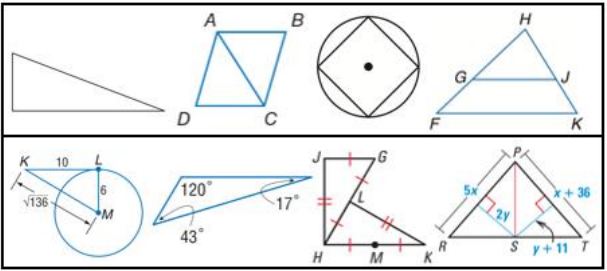

Fig. 9. Samples in Geometry3K (first line) and PGDP5K (second line).

## C. Experimental Results

We experimented with methods Freeman and GEOS on datasets Geometry3K and PGDP5K. The detection performance is shown in Table III. It is obvious that the existing geometric primitive extraction methods do not perform well on all datasets, particularly on point and line. Compared with the existing datasets, these methods even have lower performance on PGDP5K due to its complex layouts and various scale. These results indicate that there is a large room to improve for geometric primitive extraction methods.

For a more advanced evaluation to our dataset, we conducted experiments with the state-of-the-art model Inter-GPS in GDL generation. Table IV shows the experimental results of all, geo2geo and non-geo2geo relations, where IMP-Geometry3K is the re-annotated dataset of Geometry3K using our annotation method. It is shown that the results on IMP-Geometry3K have been evidently improved compared with those on Geometry3K. On the other hand, however, by the state-of-the-art method Inter-GPS, the performance on PGDP5K is significantly lower than that on IMP-Geometry3K, in that the F1 value of all relationships on PGDP5K is only 66.07%.

To further highlight the contributions of our dataset and fine-grained annotation, we visualize some samples and their parsing results by Inter-GPS. Firstly, our fine-grained annotation improves the performance of diagram parsing model. As shown in Figure 10, after re-annotation of Geometry3K using the proposed fine-grained annotation method, some relations ignored previously can be identified by Inter-GPS. Some primitives such as tangent points still cannot be detected correctly, however. Secondly, our dataset is more realistic and challenging than previous datasets as shown in Figure 9. Compared with diagrams in Geometry3K, the diagrams in our dataset are more variable in structures and primitive relationships, so that even the state-of-the-art model Inter-GPS performs poorly on them.

TABLE IV. EVALUATION RESULTS OF GEOMETRIC DESCRIPTION LANGUAGE GENERATION.

|  |  | Geometry3K | IMP-Geometry3K | PGDP5K |
|---|---|---|---|---|
| All | P | 64.30 | 67.71 | 61.06 |
|  | R | 74.06 | 76.39 | 71.97 |
|  | F1 | 68.83 | 71.79 | 66.07 |
|  | LS | 70.22 | 73.71 | 65.70 |
|  | AS | 48.92 | 50.08 | 44.40 |
|  | PR | 44.59 | 45.26 | 40.00 |
|  | TS | 34.78 | 34.28 | 27.30 |
| Geo2Geo | P | \ | 69.64 | 61.66 |
|  | R | \ | 84.93 | 86.09 |
|  | F1 | \ | 76.53 | 71.86 |
|  | LS | \ | 69.88 | 63.90 |
|  | AS | \ | 56.24 | 49.40 |
|  | PR | \ | 74.71 | 78.70 |
|  | TS | \ | 47.59 | 40.80 |
| Non-geo2Geo | P | \ | 75.38 | 66.99 |
|  | R | \ | 71.03 | 64.88 |
|  | F1 | \ | 73.14 | 65.92 |
|  | LS | \ | 77.04 | 67.30 |
|  | AS | \ | 59.07 | 49.80 |
|  | PR | \ | 50.92 | 45.70 |
|  | TS | \ | 48.59 | 40.50 |

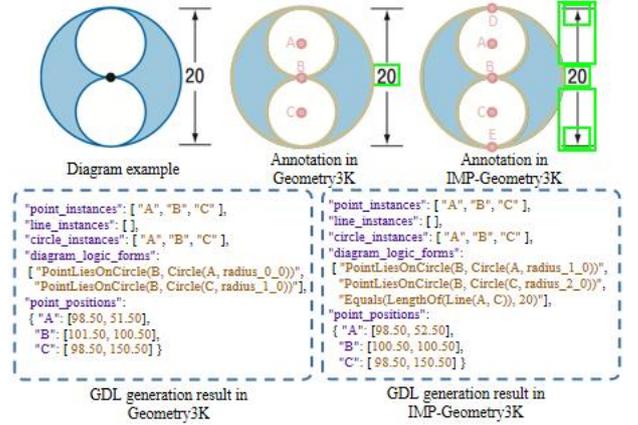

Fig. 10. Visualization of experimental results by Inter-GPS.

## V. CONCLUSION AND PROSPECT

In this paper, we propose a large dataset for plane geometry diagram parsing (PGDP), which is an important problem in document image understanding and intelligent education. The proposed dataset overcomes the insufficiencies of existing datasets in scale and fine-grained annotation, so as to facilitate research of PGDP in deep learning era. We performed experiments using existing parsing methods on the proposed PGDP5K dataset and a re-annotated dataset IMP-Geomerty3K from an existing one. The results reveal that the proposed dataset is challenging and there is still much room for improvement in the field of geometry diagram parsing.

Based on the proposed dataset, we are developing deep learning parsing algorithms for extracting primitives and relationships. Geometry problem solving is also being considers, which requires the semantic combination of geometry diagram and problem text, and the geometry diagram dataset needs to be enhanced with the problem text.